\documentclass{ieeeaccess}
\usepackage{cite}
\usepackage{amsmath,amssymb,amsfonts}
\usepackage{algorithmic}
\usepackage{graphicx}
\usepackage{textcomp}
\usepackage{amsmath}
\usepackage{multirow}
\usepackage{support-caption}
\usepackage{subcaption}
\usepackage{makecell}
\usepackage{placeins}
\def\BibTeX{{\rm B\kern-.05em{\sc i\kern-.025em b}\kern-.08em
    T\kern-.1667em\lower.7ex\hbox{E}\kern-.125emX}}
\begin{document}
\history{Received 25 January 2023, accepted 3 February 2023, date of publication 10 February 2023, date of current version 15 February 2023.
}
\doi{10.1109/ACCESS.2023.3243829}

\title{Self-Supervised Image Denoising for Real-World Images with Context-aware Transformer}
\author{\uppercase{Dan Zhang}\authorrefmark{1}, 
\uppercase{Fangfang Zhou\authorrefmark{2}}}
\address[1]{Senslab Technology Co., Ltd, PuDong New Area, Shanghai, 201203, China (e-mail: zhangdan\underline{ }fiona@163.com)}
\address[2]{Senslab Technology Co., Ltd, PuDong New Area, Shanghai, 201203, China  (e-mail: 51174700059@stu.ecnu.edu.cn)}

\markboth
{Author \headeretal: Preparation of Papers for IEEE TRANSACTIONS and JOURNALS}
{Author \headeretal: Preparation of Papers for IEEE TRANSACTIONS and JOURNALS}

\corresp{Corresponding author: Dan Zhang (e-mail: zhangdan\underline{ }fiona@163.com).}

\begin{abstract}
In recent years, the development of deep learning has been pushing image denoising to a new level. Among them, self-supervised denoising is increasingly popular because it does not require any prior knowledge. Most of the existing self-supervised methods are based on convolutional neural networks (CNN), which are restricted by the locality of the receptive field and would cause color shifts or textures loss. In this paper, we propose a novel Denoise Transformer for real-world image denoising, which is mainly constructed with Context-aware Denoise Transformer (CADT) units and Secondary Noise Extractor (SNE) block. CADT is designed as a dual-branch structure, where the global branch uses a window-based Transformer encoder to extract the global information, while the local branch focuses on the extraction of local features with small receptive field. By incorporating CADT as basic components, we build a hierarchical network to directly learn the noise distribution information through residual learning and obtain the first stage denoised output. Then, we design SNE in low computation for secondary global noise extraction. Finally the blind spots are collected from the Denoise Transformer output and reconstructed, forming the final denoised image. Extensive experiments on the real-world SIDD benchmark achieve 50.62/0.990 for PSNR/SSIM, which is competitive with the current state-of-the-art method and only 0.17/0.001 lower. Visual comparisons on public sRGB, Raw-RGB and greyscale datasets prove that our proposed Denoise Transformer has a competitive performance, especially on blurred textures and low-light images, without using  additional knowledge, e.g., noise level or noise type, regarding the underlying unknown noise.
\end{abstract}


\titlepgskip=-15pt

\maketitle

\section{Introduction}
\label{sec:introduction}
During the image acquisition process of image sensors, CCD and CMOS, various noises are introduced due to the influence of sensor material properties, working environment, electronic components and circuit structure.In addition, due to the imperfection of transmission media and recording equipment, digital images are often attacked by various noises. There are basically four types of common noise in images: Gaussian noise, Poisson noise, multiplicative noise, and salt and pepper noise. Image denoising is an inevitable step in image processing, and its denoising effect has a huge impact on the subsequent image processing process. Traditional image denoising algorithms \cite{b1, b2} are slow and less robust. With the development of deep learning, image denoising algorithms have made great progress. Although some progress has been made in traditional methods \cite{b3}, supervised denoising models \cite{b4, b5, b6, b7} have relatively better denoising effects on public datasets.

However, supervised image denoising requires noisy-clean data pairs, which are very difficult to obtain in practical applications. The most common approach is to add Additive White Gaussian Noise (AWGN) or other simulated real-world noise to a clean image and artificially synthesise a noisy image to form noisy-clean pairs \cite{b4, b7, b8, b9, b10}. However, there is an unavoidable gap between the noise synthesised by noise modelling and the real-world noise. Therefore, the denoising performance of this synthesised type of denoising model will be greatly reduced when denoising real-world images.

Under these circumstances, many self-supervised training methods \cite{b11}--\cite{b16} have emerged that do not require clean images. Noise2Noise \cite{b16} trains the model with only two noisy images and achieves  denoising performance comparable to other supervised algorithms. However, it requires two fully aligned noisy images, which are also difficult to obtain in practical applications. Then Noiser2noise \cite{b17}, NAC \cite{b18} are proposed to add the same type of noise to the existing noisy image to form a noiser-noise pair, which means that they need to know the exact type of noise in the image as a prior. IDR \cite{b12} adopts an iterative approach, taking the noisy images as inputs to the existing denoising model trained with noiser-noise pairs and treating the trained output as the next round of optimisation targets to further refine the denoising model. In this way, the denoising model is optimised by iteration, which can easily cause the final denoised image to be unduly over-smoothened. Noise2Void \cite{b14} proposes a blind spot network (BSN) denoising method based on the assumption that pixel signals in the image are spatial correlated, while noise signals are spatially independent with zero mean. In recent years, many publications \cite{b14, b19, b20, b21} have proved that BSN is very effective in denoising synthetic noise. However, noise in real-world is usually spatially continuous. In order to break the spatial connection of noise, AP-BSN \cite{b22} performs 5 times pixel-shuffle downsampling (PD) on the input before training. What’s more, AP-BSN \cite{b22} adoptes centre-masked convolution kernel and dilated convolution layer (DCL) to obtain the effect of blind spots during forward propagation. However, if the step of PD is too large, it will cause irreparable damage to the spatial information, but if it is too small, it will not be able to break the spatial connection of the noise \cite{b22}.  

Most of state-of-the-art BSN denoising methods \cite{b13, b14, b15, b20, b21, b22} ultilize CNN-based networks, which are restricted by the locality of the receptive field. Therefore, they are difficult to denoise in the presence of high noise or noisy blurred textures, while repeated training leads to over-smoothing of images. In this case, we consider it is comprehensive to combine local features with the global information extracted by Transformer. The addition of global information efficiently compensates for the defects of local receptive field extraction features. In this paper, we propose the Context-aware Denoise Transformer for self-supervised denoising in two stages. For the first stage, we use a hierarchical CADTs to directly extract noise information, which will be subtracted from the original image to obtain the first denoised result. For the second stage, we use the SNE module to extract the global noise on the first denoised result and the final denoised result can be obtained by residual learning. Extensive experiments demonstrate the effectiveness and superiority of our Denoise Transformer.

The main contributions of our work are as follows:

1. We are the first to propose a Transformer-based network suitable for self-supervised image denoising.

2. We propose a new structural unit CADT that can fuse global information and local features.

3. We add a secondary global noise extractor to realise two stage denoising.

4. Our Denoise Transformer is competitive with current state-of-the-art methods in self-supervised real-world image denoising.

\section{Related Work}
\subsection{Supervised image denoising based on CNN}
Zhang et al. \cite{b4} first proposed image denoising based on deep learning. They proposed DnCNN trained with generated noisy-clean pairs by artificially adding AWGN to clean images. Then, U-Net \cite{b23} became a more widely used denoising baseline with a characteristic of multi-scale features. Afterwards, many publications proposed deep learning image denoising methods based on adding AWGN to the noisy image \cite{b5, b6, b8, b9, b10, b24, b25, b26, b27}. However, there exists a significant gap between the artificially added AWGN and the real-world noise, and these methods are not ideal for denoising in real-world applications. Several publications \cite{b28, b29} added Poisson noise corresponding to shot noise and Gaussian noise corresponding to read noise, to Raw-RGB images. After denoising in Raw-RGB space, the final denoised result image was converted back to sRGB space using ISP tools. For this denoising method, accurate noise estimation and modelling are essential for success. Although the noise obtained by statistical modelling reduced the gap between the synthetic noise and the real noise, the injected noise was not real after all. This method can mitigate the performance degradation of the denoising model in practical applications, but not enough to eliminate it. The mismatch between the training process and the application deployment limits the practical application of this method. To solve this problem perfectly, it is undoubtedly that a direct and effective method is to use the noisy-clean pairs in the real world \cite{b30, b31}. However, generating such noisy-clean pairs requires massive human labour and cost, which makes it impractical.
 
\subsection{Self-supervised image denoising methods}
Noise2Noise \cite{b16} used two perfectly aligned noisy images taken in the same scene as the input and the target respectively. When training the denoising model , L2 loss was used to minimise the difference between the noise-noise pairs, making the model capable of denoising. Then, Noise2void \cite{b14}, Noise2self \cite{b13}, Probabilistic noise2void \cite{b15}, Neigbor2Neigbor \cite{b11}, IDR \cite{b12}, CVF-SID \cite{b32}, Blind2Unblind \cite{b33}, and AP-BSN \cite{b22} were proposed to use only noisy images for training, instead of noisy-clean pairs. Neighbor2Neighbor \cite{b11} directly selected two adjacent pixels in 2*2 neighbourhood within a single Raw-RGB image at random to synthesise one sub-noise image, and the remaining pixels formed another sub-noise image. Two sub-noise images obtained in this way formed a noise-noise pair, and one of them would participate in backpropagation during training. However, training with only sub-images would inevitably lose some detail. Noise2void tried to take the whole image into account. Noise2void took the noisy image with the masked pixels as input, and the complete noisy image was regarded as the target. In this way, the masked pixels would never be seen during training, which could easily lead to details loss or over-smoothing in the image. Blind2Unblind \cite{b33} used a global masker to generate interleaved blind spots and collected blind spots after model denoising to reconstruct the denoised output. The use of global masker solves this problem perfectly, but when the noise is large, the point-like mask used in Blind2Unblind \cite{b33} will be powerless. For spatially connected noise, AP-BSN \cite{b22} used pixel-shuffle downsampling (PD)  to break the spatial connection of the noise, and then used the masked convolution kernel to extract features at the very beginning of the model. Thus, the features of the masked patches were obtained from the surrounding pixels. The effect of blind spots could be achieved by combining DCLs with the subsequent steps corresponding to the size of the convolution kernel. Finally, the current pixel can be restored from the noise with the surrounding pixels through the inference process. However, if the noise is strongly spatially connected over a large area, PD will fail to break the spatial connection, and the pixels recovered from the surroundings should still be noise, resulting in color shifts or textures loss in the desoised results.

\subsection{Transformer denoising methods}
Transformers have achieved great success in the field of computer vision \cite{b34, b35, b36, b37, b38}. These methods are based on window-based image blocks and utilize a multi-head attention mechanism, which has outstanding advantages in capturing global image information. ViT \cite{b36} showed that pure Transformers can be applied directly to sequences of image patches without overlap, and performed well on image classification tasks. Liu et al. proposed the Swin Transformer \cite{b34}, which had a hierarchical structure where cross-window connections were captured by a shifted windowing scheme. Chen et al. developed IPT \cite{b38}, a pre-trained Transformer model for low-level computer vision tasks. Liang et al. extended Swin Transformer for image restoration and proposed SwinIR \cite{b35}, which achieved state-of-the-art performance on image super-resolution and denoising. To the best of our knowledge, the existing transformer-based image denoising methods are supervised image denoising \cite{b34, b35, b39}. Due to the global feature capturing capability of Transformer, the image features obtained based on BSNs are close to the identity transformation with revisible loss. Alternatively, the original pixel can be easily recovered from the surroundings, resulting in unsatisfactory denoising performance.

Inspired by \cite{b34, b37}, we propose a new CADT structure aimed at self-supervised denoising for real-world images. CADT contains two branches, a global branch and a local branch. These two branches can effectively combine the global information and the local textures. Taking CADT as the basic component, we construct a hierarchical CADT structure to extract noise features. Once the noise information is obtained, it will be subtracted from the original image by residual learning and we get the first stage denoised result. What's more, we propose to add a Secondary Noise Extractor (SNE) module based on the first denoised output and to extract the global noise information on it. SNE is designed using LN and MLP with low computational complexity. The final denoised result can be obtained by removing the SNE output as a residual from the first denoised result.
\section{Method}
\begin{figure*}[htbp]    
\begin{subfigure}{.28\textwidth}
  \centering
  \includegraphics[height=1.8in]{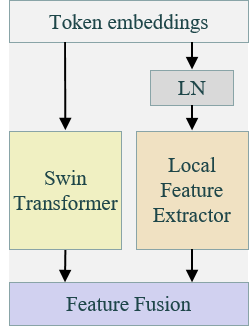}  
  \caption{CADT Framework}
  \label{fig:1-a}
\end{subfigure}
\begin{subfigure}{.52\textwidth}
  \centering
  \includegraphics[height=1.8in]{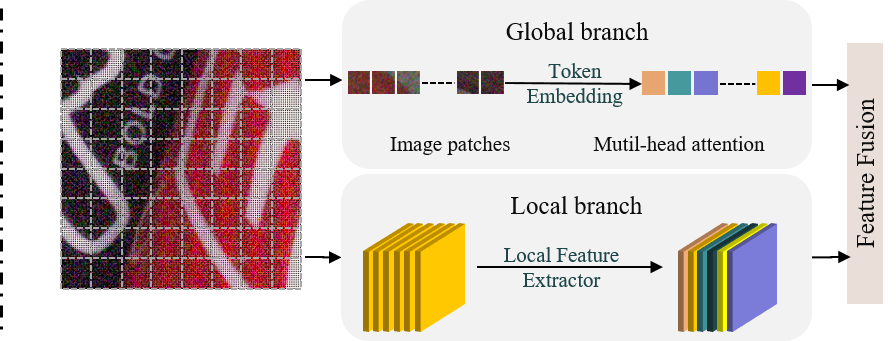}  
  \caption{CADT Workflow}
  \label{fig:1-b}
\end{subfigure}
  \setlength{\abovecaptionskip}{0cm}
  \setlength{\belowcaptionskip}{-0.6cm}
\caption{The representation of CADT. (a) shows that CADT is designed as a dual-branch structure. The left part shows the global branch of the Transformer-based method for global featrue extraction, and the right part shows the local branch, which focuses on the local feature extraction. (b) shows the workflow of CADT for feature extracting. In the global branch, token embedding is performed on image patches, and the global information is extracted by the the multi-head attention mechanism. The local branch is composed of several convolutions with small receptive fields. In particular, the deformable convolutions are used to extract the details of the local information mutation, which helps to preserve denoised image details.} 
\label{fig:figure 1} 
\end{figure*}
\subsection{CADT}
Since the pure Transformer will lead to poor  self-supervised image denoising effect, we propose a new dual-branch Transformer structure, Context-aware Denoise Transformer (CADT), as shown in Fig. \ref{fig:figure 1}. CADT includes a global feature extraction branch and a local feature extraction branch, and the framework is shown in Fig. \ref{fig:1-a}, and the workflow is shown in Fig. \ref{fig:1-b}.

\begin{figure}[t]
  \setlength{\abovecaptionskip}{0cm}
  \setlength{\belowcaptionskip}{-0.6cm}
  \centering
  \includegraphics[width=1\linewidth]{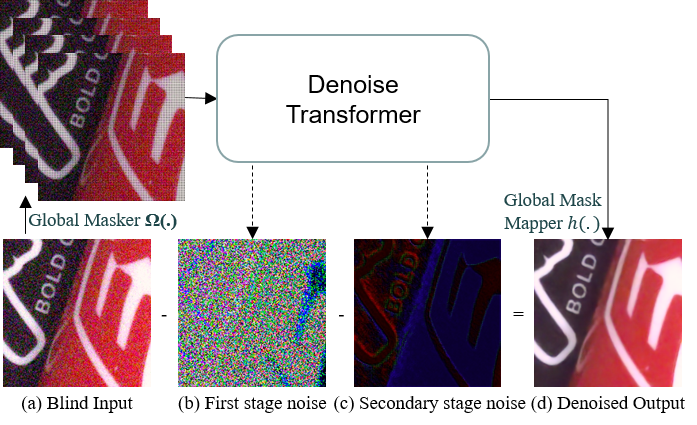}
  \caption{The framework of Denoise Transformer.}
  \label{fig:figure 2}
\end{figure}

\begin{figure*}[t]
  \setlength{\abovecaptionskip}{0cm}
  \setlength{\belowcaptionskip}{-0.6cm}
  \centering
  \includegraphics[width=1\linewidth]{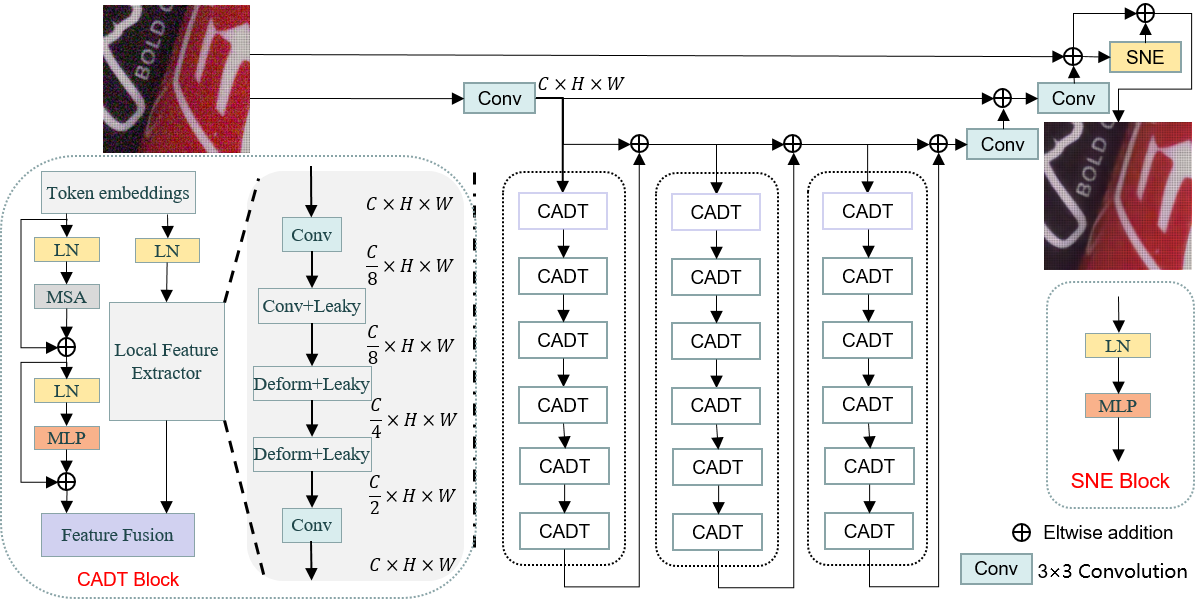}
  \caption{The architecture of Denoise Transformer.}
  \label{fig:figure 3}
\end{figure*}
{\bf Global Transformer Encoder} For the global branch, we use a window-based multi-head Transformer encoder in Swin Transformer \cite{b34} to extract the global information. The Transformer encoder contains a multi-head self-attention (MSA) module and a multi-layer perception (MLP) module. A LayerNorm (LN) layer is applied before each MSA module and each MLP, and a residual connection is applied after each module.

Assuming that the token embedding input is $E\in R^{H\times W \times C}$, the global self-attention computation can be formulated as the following Eq. (\ref{eq:eq1}):
\begin{equation}
\setlength{\abovedisplayskip}{1pt}
\setlength{\belowdisplayskip}{1pt}
\begin{split}
E = {\rm {MSA}}({\rm {LN}}(E))+E\\
{\rm {CADT}}_{{\rm {global}}}={\rm {MLP}}({\rm {LN}}(E))+E
\end{split}
\label{eq:eq1}
\end{equation}
Where LN denotes the LayerNorm, ${\rm {CADT}}_{{\rm {global}}}$ denotes the global feature extracted using Transformer Encoder.

{\bf Local Feature Extractor} For the local branch, we design a local feature extractor (LFE) to capture the local context  ${\rm {CADT}}_{\rm {local}}$, where the features of adjacent pixels and features of cross-channels are fused. The local feature extraction can be formulated as Eq. (\ref{eq:eq2}):
\begin{equation}
\setlength{\abovedisplayskip}{1pt}
\setlength{\belowdisplayskip}{1pt}
{\rm {CADT}}_{{\rm {local}}}={\rm {LFE}}({\rm {LN}}(E))
\label{eq:eq2}
\end{equation}
For the token embedding vector $E$, we normalize it by the LN layer and reshape it into a feature vector with shape of $N\times H\times W\times C$. Then we use a set of convolutions to extract and output a feature vector shaped in $N\times H\times W\times C/8$ for dimensionality reduction. We then use another ordinary convolution to undertake the feature after dimension reduction, followed by two convolutions with deformable kernels \cite{b40} to fuse contextual details with large changes and cross-channel features, which is a key design to preserve image textures during denoising. Now we have two vectors whose shapes are $N\times H\times W\times C/4$ and $N\times H\times W\times C/2$ respectively. The final shape is expanded from $N\times H\times W\times C/2$ to $N\times H\times W\times C$ by another convolution. Each convolution, except the first and the last, is followed by a LeakyReLU activation layer to improve feature selection. We summarize each computational step into the formulas as Eq. (\ref{eq:eq3}):
\begin{equation}
\begin{aligned}
f_{{\rm {reduction}}} &= {\rm {Conv}}({\rm {LN}}(E))\\
f_{{\rm {local}}}  &= {\rm {Leaky}}(D({\rm {Leaky}}(D(f_{{\rm {reduction}}})))\\
f_{{\rm {expansion}}}&= {\rm {Conv}}(f_{{\rm {local}}})\\
{\rm {CADT}}_{{\rm {local}}} &= f_{{\rm {expansion}}}
\end{aligned}
\label{eq:eq3}
\end{equation}
Finally, element-wise addition is used to fuse global and local information, effectively reducing additional parameters compared to linear or convolutional layers. The local branch adds local features to the Transformer Encoder, changing the pure transformer mapping relationship, and preserving local mutation information and texture detail.

\subsection{SNE Block}
We propose a secondary noise extraction module with low computation at the end of the Denoise Transformer. We want to further distinguish true noise from textures in a relatively clean image. And this is easier to realize on a relatively clean image than on a noisy image. Therefore, we design a low computational global information extraction mechanism inspired by the multi-layer perception mechanism. First, the denoised image is reshaped into $N\times C\times (H\times W)$ and LayerNorm operation is performed to normalize the global information of each channel. Multi-layer perception is used to exchange information between channels. With residual learning, the output of SNE would be subtracted from the first stage denoised image. That is to say, SNE learns the second noise information mapping using the first stage denoised image and the final image. Although SNE has a relatively simple design, it achieves further noise reduction while preserving image detail. The module works as a plug-in and can be portable if required. The SNE block is calculated as follows Eq. (\ref{eq:eq4}):
\begin{equation}
\setlength{\abovedisplayskip}{1pt}
\setlength{\belowdisplayskip}{1pt}
f_{{\rm {SNE}}} = {\rm {MLP}}({\rm {LN}}(x))\\
\label{eq:eq4}
\end{equation}
Where, $x$ denotes the denoised image. LN denotes the LayerNorm, and MLP denotes the multi-layer perception.
\subsection{A Choice of Loss}
With the baseline of Transformer, the local information extraction will be severely compromised if we follow \cite{b11, b33} to pass the original images through the network in a non-gradient manner. When all the noisy images and the corresponding denoised results are included in the total loss calculation, the global feature captured by Transformer encoder can easily recover the noise information of the original image. It seems that transformer can recover what it sees before, even though the pixels do not contribute to the gradient descent. To avoid this kind of the identity transformation, we do not compute the invisible loss. During the training process, only the images after the global masker are used as training objects, and the $L_2$ loss is performed using the blind input and the corresponding output. The loss function can be formulated as the following Eq. (\ref{eq:eq5}):
\begin{equation}
\begin{aligned}
\setlength{\abovedisplayskip}{1pt}
\setlength{\belowdisplayskip}{1pt}
L_2=({\rm {Noisy}}-{\rm {denoised}})^2
\end{aligned}
\label{eq:eq5}
\end{equation}
\subsection{Overall Architecture of Denoise Transformer}
Based on the CADT structure and the SNE block introduced above, we propose the Denoise Transformer framework, as shown in Fig. \ref{fig:figure 2}. The detailed architecture is shown in Fig. \ref{fig:figure 3}.

Firstly, we perform PD on the original image to break the spatial noise connection \cite{b22}, and then we use the Global-aware Mask Mapper \cite{b33} to create the blind spot effect. The mask mapper performs global denoising on the blind spots. This mechanism constrains all pixels, promotes information exchange across whole masked regions, and improves denoising performance. In this paper, the mask width is set to 4, which means that one noisy image will be mapped into 16 blind sub-images. The blind sub-images are sent to the network as inputs. We do the feature dimension mapping of the blind input with the patch embedding dimension through a convolution layer. And then the features go through hierarchical CADTs. In this paper the number of CADT groups is set to 3, each of which contains 6 CADT units for feature extraction, and a residual connection is applied after each CADT group. Next, the output of the hierarchical CADTs goes through another convolution layer, which maps the feature dimension back to the input dimension, forming the residual noise. We subtract the noise pattern from the blind image and obtain a denoised image. The denoised image then passes through SNE, and the residual secondary noise features are extracted. Finally, the denoised image is obtained by subtracting the SNE output and is subjected to the reverse operation of the Global-aware Mask Mapper to form the final denoised result.
Ablation experiments for several modules are presented in detail in Section \ref{sec:ablations}.

\section{Experiments}
\subsection{Implementation Details}
\begin{figure}[t]
  \setlength{\abovecaptionskip}{0cm}
  \setlength{\belowcaptionskip}{-0.6cm}
  \centering
  \includegraphics[width=1\linewidth]{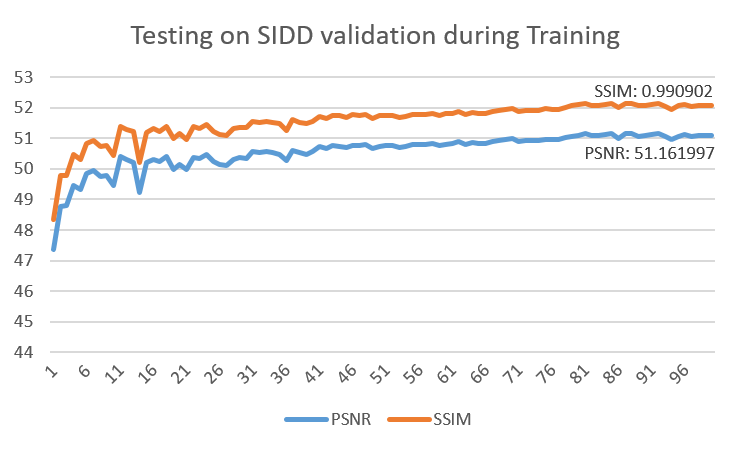}
  \caption{Denoise Transformer test results for SIDD validation in different epochs, with maximum  PSNR and SSIM marked on the curve.}
  \label{fig:figure 4}
\end{figure}

{\bf Training Details.} All models are trained with most of the same settings. Although our model aimed at real-world denoising, we perform real-world deoising experiments on real-world Raw-RGB images and greyscale images, as well as synthetic sRGB with Gaussian noise, for a more comprehensive comparison with state-of-the-art models. The detailed network is shown in Fig. \ref{fig:figure 2} and Fig. \ref{fig:figure 3}, where the token embedding dimension {$C$} is set to 60 and all convolutions used in the network have a kernel size of 3$\times$3 and a stride of 1. The batch size is set to 4 and the masker step is set to 4. We use the Adam optimizer with a weight decay of 1e-8, an initial learning rate of 0.0003 for synthetic denoising experiments in sRGB space and 0.0001 for real-world denoising experiments in Raw-RGB space and fluorescence microscopy (FM). The learning rate is multiplied by 0.25 every 20 epochs, for 100 training epochs. The images are randomly cropped into 128 × 128 patches, masked with a step size of 4, and entered into the model for training after random rotation within 90°and random horizontal or vertical flipping. All models are trained on python3.8.0, pytorch1.12.0 and Nvidia Tesla T4 GPUs. Fig. \ref{fig:figure 4} shows a plotted curve for our proposed model testing on SIDD validation in different epochs, proving the effectiveness of our network structure.

{\bf Datasets for Synthetic Denoising.} We perform synthetic experiments with varied Gaussian noise,  considering a varied Gaussian noise distribution with $\sigma\in [5, 50]$ in sRGB space are considered. We follow the settings in \cite{b11, b33} and select 44,328 images from ImageNet \cite{b41} validation set with resolutions ranging from 256×256 to 512×512, as the clean set. For the test sets, we follow \cite{b33} to repeat Kodak \cite{b42}, BSD300-test \cite{b43} and Set14 \cite{b44} by 10, 3 and 20 times, respectively, and calculate the average PSNR/SSIM.  

{\bf Datasets for real-world Denoising.} For real-world denoising in Raw-RGB space, we take the public dataset of SIDD for experiments. The SIDD Medium dataset contains 320 pairs of noisy-clean images. We take only Raw-RGB images in SIDD Medium \cite{b45} dataset as train set, and we take Raw-RGB images in SIDD Validation and Benchmark datasets as the valid set and test set respectively. For real-world denoising on FM with greyscale images, we use FMDD \cite{b46} dataset for experiments. FMDD contains real fluorescence microscopy images and 60,000 noisy images with different noise levels, which are obtained with commercial confocal, two-photon, and widefield microscopes and representative biological samples such as cells, zebrafish, and mouse brain tissue. Following the setting in \cite{b33}, we train on three datasets including Confocal Fish, Confocal Mice and Two Photon Mice, considering each set as both the training set and validation set at the same time and use the 50th view for visual testing.

{\bf Details of Experiments.} We use PSNR and SSIM as evaluation metrics. PSNR/SSIM on SIDD validation set is obtained by using python toolkit based on clean images and denoised images. And PSNR/SSIM on SIDD benchmark is reported from the official website through submitting our denoised result.

For fair comparison, we follow the experimental settings of Blind2Unblind \cite{b33}. We compare our proposed Denoising Transformer method with two supervised denoising methods (N2C \cite{b23} and N2N \cite{b16}) for the baseline. We also quantitatively compare the Denosing Transformer method with a traditional approach (BM3D \cite{b1}) and several self-supervised denoising algorithms (Laine19 \cite{b19}, Self2self \cite{b47}, DBSN \cite{b21}, N2V \cite{b14}, R2R \cite{b48}, Noisier2noise \cite{b17}, NBR2NBR \cite{b11} and Blind2Unblind \cite{b33}).

For the estimation of synthetic denoising, we use pre-trained models provided by \cite{b19} for N2C, N2N, and Laine19 \cite{b19}, keeping the same network architecture as \cite{b11, b19}. And then, we use a multi-channel version of BM3D, namely CBM3D \cite{b49}, to denoise Gaussian noise using the $\sigma$ noise level estimation method in \cite{b50}. For Self2self, Noisier2noise, DBSN, R2R, NBR2NBR and Blind2Unblind, the official implementations are used for experiments. Since Possion noise can be transformed into Gaussian distribution \cite{b51}, we will not show the denoising performance specifically for Possion noise.  

For estimating of real-world denoising in Raw-RGB space, all the methods we use are based on their official implementations, trained on the SIDD Medium dataset and well shown through ISP tools \cite{b45}. For BM3D, we use BM3D-CFA \cite{b52} for Raw-RGB denoising. For the learning-based networks, we split the single channel raw image into four sub-images following the Bayer pattern. We gather four sub-images to form a four-channel image for denoising and then reconstruct the Raw-RGB space from the denoised image. For the estimation of the real-world denoising on FM greyscale set, all compared methods are retrained using the corresponding authors' implementation.

\begin{figure*}[htbp]
  \setlength{\abovecaptionskip}{0cm}
	\centering
		\begin{subfigure}{0.16\linewidth}
		\setlength{\abovecaptionskip}{1pt}
		\centering
		\includegraphics[width=1.11in]{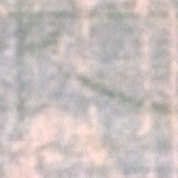}
	\end{subfigure}
	\begin{subfigure}{0.16\linewidth}
		\setlength{\abovecaptionskip}{1pt}
		\centering
		\includegraphics[width=1.11in]{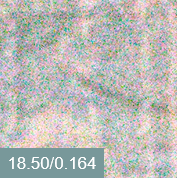}
	\end{subfigure}
	\begin{subfigure}{0.16\linewidth}
		\setlength{\abovecaptionskip}{1pt}
		\centering
		\includegraphics[width=1.11in]{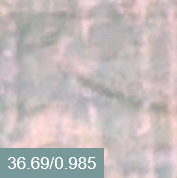}
	\end{subfigure}
	\begin{subfigure}{0.16\linewidth}
		\setlength{\abovecaptionskip}{1pt}
		\centering
		\includegraphics[width=1.11in]{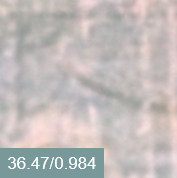}
	\end{subfigure}
	\begin{subfigure}{0.16\linewidth}
		\setlength{\abovecaptionskip}{1pt}
		\centering
		\includegraphics[width=1.11in]{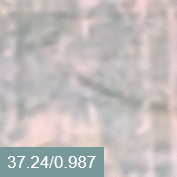}
	\end{subfigure}
	\begin{subfigure}{0.16\linewidth}
		\setlength{\abovecaptionskip}{1pt}
		\centering
		\includegraphics[width=1.11in]{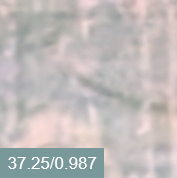}
	\end{subfigure}

\vspace{0.06cm}
\hspace{0.01cm}
	\begin{subfigure}{0.16\linewidth}
		\setlength{\abovecaptionskip}{1pt}
		\centering
		\includegraphics[width=1.11in]{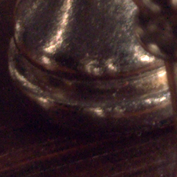}
       \captionsetup{font={scriptsize}}
		\caption{Clean}
	\end{subfigure}
	\begin{subfigure}{0.16\linewidth}
		\setlength{\abovecaptionskip}{1pt}
		\centering
		\includegraphics[width=1.11in]{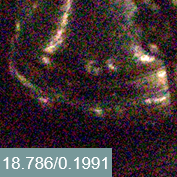}
       \captionsetup{font={scriptsize}}
		\caption{Noisy}
	\end{subfigure}
	\begin{subfigure}{0.16\linewidth}
		\setlength{\abovecaptionskip}{1pt}
		\centering
		\includegraphics[width=1.11in]{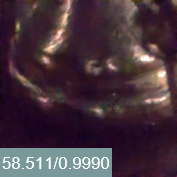}
       \captionsetup{font={scriptsize}}
		\caption{NBR2NBR}
	\end{subfigure}
	\begin{subfigure}{0.16\linewidth}
		\setlength{\abovecaptionskip}{1pt}
		\centering
		\includegraphics[width=1.11in]{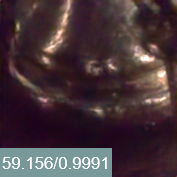}
       \captionsetup{font={scriptsize}}
		\caption{Blind2Unblind}
	\end{subfigure}
	\begin{subfigure}{0.16\linewidth}
		\setlength{\abovecaptionskip}{1pt}
		\centering
		\includegraphics[width=1.11in]{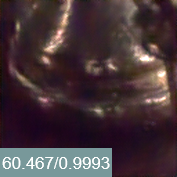}
       \captionsetup{font={scriptsize}}
		\caption{First Denoised}
	\end{subfigure}
	\begin{subfigure}{0.16\linewidth}
		\setlength{\abovecaptionskip}{1pt}
		\centering
		\includegraphics[width=1.11in]{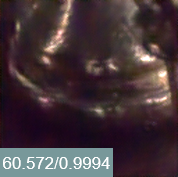}
       \captionsetup{font={scriptsize}}
		\caption{Second Denoised}
	\end{subfigure}
  \setlength{\belowcaptionskip}{-0.6cm}
  \caption{Visual Comparison of different self-supervised methods denoising for Raw-RGB images in SIDD validation.}
  \label{fig:figure 5}
\end{figure*}

\subsection{Ablation Study}
\label{sec:ablations}
In order to quantitatively evaluate the performance of different networks and training processes, several comprehensive ablation experiments are conducted based on the metric of PSNR/SSIM of real-world Raw-RGB images from SIDD validation set.

\begin{table}[htbp]
  \setlength{\abovecaptionskip}{0cm}
  \setlength{\belowcaptionskip}{0cm}
  \centering
  \caption{Quantitative comparison of various branches and their combinations performance on the SIDD validation. (1) BL: Baseline model. CADT: Dual branch composed of global and loacal extractors. SNE: Secondary noise extractor. (2) $\checkmark$ denotes using the corresponding branch. (3) The best of PSNR/SSIM is highlighted in {\bf bold}.}
  \label{tab:table 1}  
  \begin{tabular*}{\hsize}{c@{\extracolsep{\fill}}c@{\extracolsep{\fill}}c@{\extracolsep{\fill}}c@{\extracolsep{\fill}}}
    \hline
    BL & CADT & SNE & SIDD Validation \\
    \hline
    \checkmark &  &   & 46.8/0.975 \\
	 \checkmark &	 & \checkmark &47.2/0.975\\
    \checkmark & \checkmark &  & 50.8/0.990\\
    \checkmark & \checkmark &\checkmark  &  {\bf 51.16/0.991}\\
    \hline
  \end{tabular*}
\vspace{-1em}
\end{table}

{\bf Ablation on the network architecture.} For the design of the network structure, we are inspired to introduce Transformer, which has achieved good performance in computer vision, as a baseline for self-supervised denoising. Based on this, CADT component, SNE block, and the overall Denoise Transformer are successively proposed. Therefore, in the ablation experiments, we set the following variables:

-{\bf Baseline}. We use a tiny version of SwinIR \cite{b35}, which consists of pure transformer encoders.

-{\bf CADT}. This module extends the pure transformer based on the baseline, which consists of a global branch and a local branch, forming a dual-branch structure.

-{\bf SNE}. It is a secondary noise extractor. It is designed to be connected after the first stage denoised result, and to extract the global noise on the first stage denoised result to realize the secondary denoising.

-+{\bf CADT}+{\bf SNE}. The overall network architecture of our proposed Denoise Transformer.

The quantitative evaluation of the SIDD validation in Raw-RGB space is shown in Table \ref{tab:table 1}. It can be seen that direct use of pure Transformer method does not perform well in self-supervised denoising. Our possible reason is that over-extraction of global features prevents local features from being effectively denoised.  Based on  the original global branch, the addition of the local branch to CADT and SNE can both improve the self-supervised denoising ability. In particular, CADTs show the surprising performance for denoising. To show the performance of SNE, we also present a set of visual intermediate results in Fig. \ref{fig:figure 5}. From Fig. \ref{fig:figure 5} we can see that SNE can slightly contribute to denoising, which is in line with our expectations. In the ideal case, there is little noise present in the first stage denoised result and therefore we design SNE in low computation. SNE is the icing on the cake. By effectively combining it with global features and local information, the global noise estimation, denoising and local texture reconstruction can be better optimized and complement each other.

{\bf Ablation on Transformer loss.}
Based on the overall architecture of Denoise Transformer, we conduct experiments on three different losses. Re-visible Loss1 means that the forward propagation and the total loss calculation are consistent with NBR2NBR \cite{b11}. Re-visible Loss2 means that the forward propagation and the total loss calculation are consistent with Blind2Unblind \cite{b33}. The comparison are shown in Table \ref{tab:table 2}.

\begin{table}[htbp]
  \centering
  \setlength{\abovecaptionskip}{0cm}
  \setlength{\belowcaptionskip}{0cm}
  \caption{Quantitative comparison of different losses. The best of PSNR/SSIM is highlighted in {\bf bold}.} 
  \label{tab:table 2}
  \setlength{\tabcolsep}{6mm}{
  \begin{tabular}{lc@{}lc@{}}
    \hline
    Loss Type & SIDD Validation \\
    \hline
    Re-visible Loss\cite{b11} &  44.56/0.961\\
    Re-visible Loss\cite{b36} & 48.64/0.986\\
    $L_2$ Loss &	{\bf 51.16/0.991}\\
    \hline
  \end{tabular}}
\vspace{-1em}
\end{table}
It can be seen from Tabel \ref{tab:table 2} that when the Transformer branch is used in the self-supervised denoising model, its powerful global extraction ability is enhanced under the re-visible action, making the mapping result of the self-supervised denoising network onto the original noisy image close to identity transformation. This makes it easier to recover the noise information of the blind spot, which greatly reduces the denoising ability. In this paper we use $L_2$ loss instead of the re-visible loss for Denoise Transformer.

\begin{figure}[t]
  \setlength{\abovecaptionskip}{0cm}
  \setlength{\belowcaptionskip}{-0.6cm}
  \centering
  \includegraphics[width=1\linewidth]{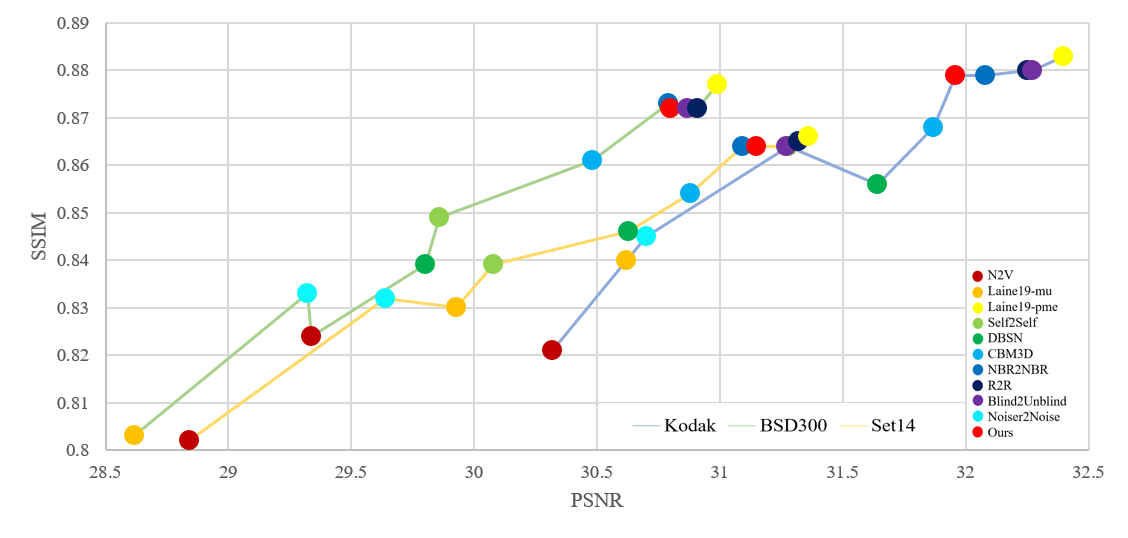}
  \caption{The curve of quantitative comparison of several self denoising methods with Gaussian noise $\sigma \in [5, 50]$.}
  \label{fig:figure 6}
\end{figure}
\begin{figure*}[htbp]
  \setlength{\abovecaptionskip}{0cm}
	\centering
	\begin{subfigure}{0.15\linewidth}
		\centering
		\includegraphics[height=0.71in]{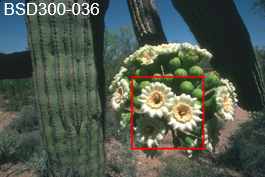}
	\end{subfigure}
	\begin{subfigure}{0.1\linewidth}
		\centering
		\includegraphics[height=0.71in]{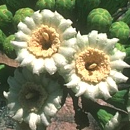}
	\end{subfigure}
	\begin{subfigure}{0.1\linewidth}
		\centering
		\includegraphics[height=0.71in]{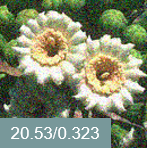}
	\end{subfigure}
	\begin{subfigure}{0.1\linewidth}
		\centering
		\includegraphics[height=0.71in]{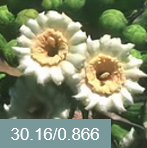}
	\end{subfigure}
	\begin{subfigure}{0.1\linewidth}
		\centering
		\includegraphics[height=0.71in]{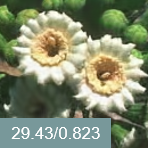}
	\end{subfigure}
	\begin{subfigure}{0.1\linewidth}
		\centering
		\includegraphics[height=0.71in]{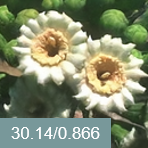}
	\end{subfigure}
	\begin{subfigure}{0.1\linewidth}
		\centering
		\includegraphics[height=0.71in]{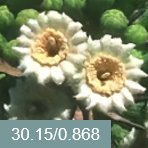}
	\end{subfigure}
	\begin{subfigure}{0.1\linewidth}
		\centering
		\includegraphics[height=0.71in]{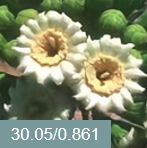}
	\end{subfigure}
	\begin{subfigure}{0.1\linewidth}
		\centering
		\includegraphics[height=0.71in]{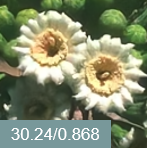}
	\end{subfigure}

 \vspace{0.06cm}
		\begin{subfigure}{0.15\linewidth}
		\centering
		\includegraphics[height=0.71in]{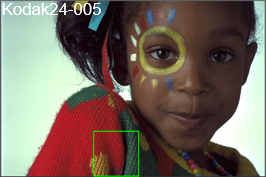}
	\end{subfigure}
	\begin{subfigure}{0.1\linewidth}
		\centering
		\includegraphics[height=0.71in]{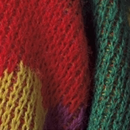}
	\end{subfigure}
	\begin{subfigure}{0.1\linewidth}
		\centering
		\includegraphics[height=0.71in]{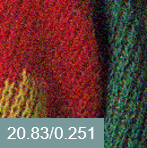}
	\end{subfigure}
	\begin{subfigure}{0.1\linewidth}
		\centering
		\includegraphics[height=0.71in]{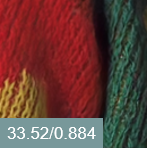}
	\end{subfigure}
	\begin{subfigure}{0.1\linewidth}
		\centering
		\includegraphics[height=0.71in]{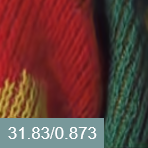}
	\end{subfigure}
	\begin{subfigure}{0.1\linewidth}
		\centering
		\includegraphics[height=0.71in]{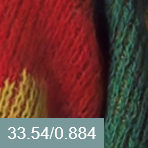}
	\end{subfigure}
	\begin{subfigure}{0.1\linewidth}
		\centering
		\includegraphics[height=0.71in]{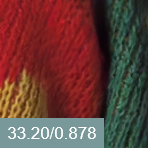}
	\end{subfigure}
	\begin{subfigure}{0.1\linewidth}
		\centering
		\includegraphics[height=0.71in]{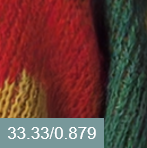}
	\end{subfigure}
	\begin{subfigure}{0.1\linewidth}
		\centering
		\includegraphics[height=0.71in]{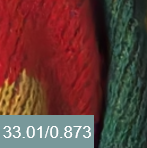}
	\end{subfigure}

 \vspace{0.06cm}
 \hspace{0.005cm}
		\begin{subfigure}{0.15\linewidth}
		\setlength{\abovecaptionskip}{1pt}
		\centering
		\includegraphics[height=0.709in]{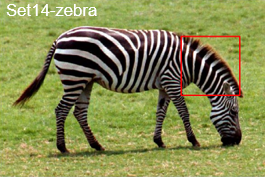}
       \captionsetup{font={scriptsize}}
		\caption*{Original}
	\end{subfigure}
	\begin{subfigure}{0.1\linewidth}
		\setlength{\abovecaptionskip}{1pt}
		\centering
		\includegraphics[width=0.709in,height=0.709in]{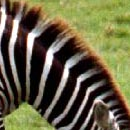}
       \captionsetup{font={scriptsize}}
		\caption*{Clean}
	\end{subfigure}
	\begin{subfigure}{0.1\linewidth}
		\setlength{\abovecaptionskip}{1pt}
		\centering
		\includegraphics[width=0.706in,height=0.709in]{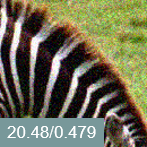}
       \captionsetup{font={scriptsize}}
		\caption*{Noisy}
	\end{subfigure}
	\begin{subfigure}{0.1\linewidth}
		\setlength{\abovecaptionskip}{1pt}
		\centering
		\includegraphics[width=0.706in,height=0.709in]{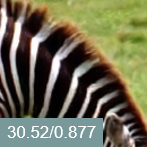}
       \captionsetup{font={scriptsize}}
		\caption*{Baseline, N2N}
	\end{subfigure}
	\begin{subfigure}{0.1\linewidth}
		\setlength{\abovecaptionskip}{1pt}
		\centering
		\includegraphics[width=0.709in,height=0.709in]{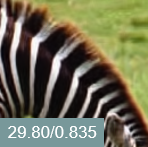}
       \captionsetup{font={scriptsize}}
		\caption*{CBM3D}
	\end{subfigure}
	\begin{subfigure}{0.1\linewidth}
		\setlength{\abovecaptionskip}{1pt}
		\centering
		\includegraphics[width=0.709in,height=0.709in]{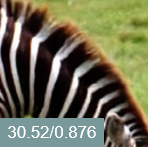}
       \captionsetup{font={scriptsize}}
		\caption*{Laine19-pme}
	\end{subfigure}
	\begin{subfigure}{0.1\linewidth}
		\setlength{\abovecaptionskip}{1pt}
		\centering
		\includegraphics[width=0.709in,height=0.709in]{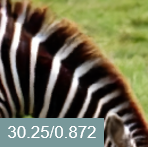}
       \captionsetup{font={scriptsize}}
		\caption*{NBR2NBR}
	\end{subfigure}
	\begin{subfigure}{0.1\linewidth}
		\setlength{\abovecaptionskip}{1pt}
		\centering
		\includegraphics[width=0.709in,height=0.709in]{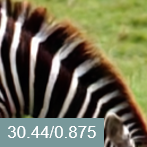}
       \captionsetup{font={scriptsize}}
		\caption*{Blind2Unblind}
	\end{subfigure}
	\begin{subfigure}{0.1\linewidth}
		\setlength{\abovecaptionskip}{1pt}
		\centering
		\includegraphics[width=0.709in,height=0.709in]{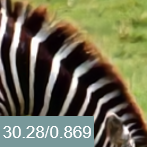}
       \captionsetup{font={scriptsize}}
		\caption*{Ours}
	\end{subfigure}
  \setlength{\belowcaptionskip}{-0.6cm}
  \caption{Visual Comparison of different methods denoising for images in sRGB space with Gaussion noise $\sigma=25$.}
  \label{fig:figure 7}
\end{figure*}
\subsection{Comparisons with State-of-the-Art}
{\bf Results for Synthetic Denoising.} For the widely used Gaussian noise, the quantitative comparison in synthetic denoising is shown in Fig. \ref{fig:figure 6}. To better evaluate the denosing ability of state-of-the-art models, we have varied the Gaussian noise in a wide range $\sigma \in [5, 50]$. Fig. \ref{fig:figure 6} shows that our method outperforms the traditional denoising method BM3D and several self-supervised denoising methods, such as Self2self, Laine19-mu, DBSN and NBR2NBR. Fig. \ref{fig:figure 7} shows the visual comparison of the denoised images injected with Gaussian noise with noise level $\sigma=25$. Denoise Transformer achieves competitive visual quality among traditional and self-supervised CNN denoisers. What's more, its performance is close to the SOTA of Blind2Unblind \cite{b34} and even surpasses it on single images with low saturated textures, proving the advantage of preserving image textures of our proposed method. However, Fig. \ref{fig:figure 7} also shows that Denoise Transformer does not perform as well as Blind2Unblind when processing images with high saturation. The possible reason is that in the high saturation region, the extracted global features  are at a high level while the local details have relatively small changes. And the small changes in texture would be considered as noise and removed.

{\bf Results for real-world Denoising.} For real-world denoising, Table \ref{tab:table 3} quantitatively compares the denoising performance of different methods on Raw-RGB space of SIDD benchmark and validation, and Table \ref{tab:table 4} quantitatively compares the denoising performance on FM dataset. Our proposed Denoise Transformer outperforms the traditional method BM3D and Laine19 to a large extent, and it even outperforms Neighbor2Neighbor by 0.15dB and 0.10dB for SIDD benchmark and validation, and it outperforms NBR2NBR more on FMDD dataset.

 \FloatBarrier
\begin{table}[htbp]
  \setlength{\abovecaptionskip}{0cm}
  \setlength{\belowcaptionskip}{0cm}
  \centering
   \caption{Quantitative comparison of denoising methods on synthetic datasets in Raw-RGB space of SIDD benchmark. DT denotes our Denoised Transformer. The highest PSNR/SSIM  is highlighted in {\bf bold}, while the second is \underline {underlined}.}
  \label{tab:table 3}
  \begin{tabular*}{\hsize}{l@{\extracolsep{\fill}}l@{\extracolsep{\fill}}c@{\extracolsep{\fill}}c@{\extracolsep{\fill}}}
    \hline
    Methods & Network & \makecell[c]{SIDD \\ Benchmark} & \makecell[c]{SIDD \\ Validation} \\
	\hline
	Baseline, N2C\cite{b23}	&U-Net \cite{b23}	&50.60/0.991	&51.19/0.991\\
	Baseline, N2N\cite{b16}	&U-Net \cite{b23}	&50.62/0.991	&51.21/0.991\\
	\hline
	BM3D \cite{b1}	&-	&48.60/0.986	&48.92/0.986\\
	N2V \cite{b14}	&U-Net \cite{b23}	&48.01/0.983	&48.55/0.984\\
	Laine19-mu (G) \cite{b19}	&U-Net \cite{b23}	&49.82/0.989	&50.44/0.990\\
	Laine19-pme (G) \cite{b19}	&U-Net \cite{b23}	&42.17/0.935	&42.87/0.939\\
	Laine19-mu (P) \cite{b19}	&U-Net \cite{b23}	&50.28/0.989	&50.89/0.990\\
	Laine19-pme (P) \cite{b19}	&U-Net \cite{b23}	&48.46/0.984	&48.98/0.985\\
	DBSN \cite{b21}	&DBSN \cite{b21} &49.56/0.987	&50.13/0.988\\
	R2R \cite{b48}	&U-Net \cite{b23}	&46.70/0.978	&47.20/0.980\\
	NBR2NBR \cite{b11}	&U-Net \cite{b23}	&50.47/\underline{0.990}	&51.06/\underline{0.991}\\
	Blind2Unblind\cite{b33}	&U-Net \cite{b23}	&{\bf 50.79/0.991}	&{\bf 51.36/0.992}\\
	Ours	&DT	&\underline{50.62/0.990}	&\underline{51.16/0.991}\\
	\hline
  \end{tabular*}
\vspace{-2em}
\end{table}

\begin{figure}[htbp]
  \setlength{\abovecaptionskip}{0cm}
  \setlength{\belowcaptionskip}{0cm}
	\centering
	\begin{subfigure}{0.41\linewidth}
  		\setlength{\abovecaptionskip}{1pt}
		\centering
		\includegraphics[height=1.375in]{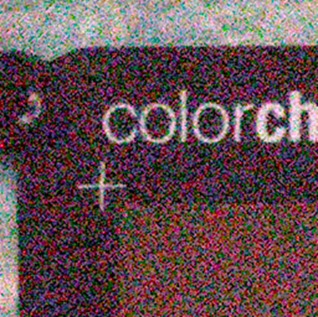}
		\captionsetup{font={scriptsize}}
		\caption*{Noisy}
	\end{subfigure}
	\begin{subfigure}{0.185\linewidth}
		\setlength{\abovecaptionskip}{1pt}
		\centering
		\includegraphics[height=0.6251in]{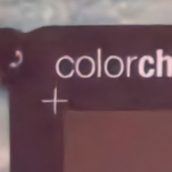}
		\captionsetup{font={scriptsize}}
		\caption*{Baseline, N2C}
		\includegraphics[height=0.6251in]{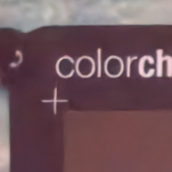}
		\captionsetup{font={scriptsize}}
		\caption*{Baseline, N2N}
	\end{subfigure}
	\begin{subfigure}{0.185\linewidth}
		\setlength{\abovecaptionskip}{1pt}
		\centering
		\includegraphics[height=0.6251in]{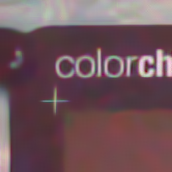}
		\captionsetup{font={scriptsize}}
		\caption*{BM3D}
		\includegraphics[height=0.6251in]{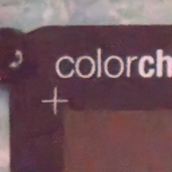}
		\captionsetup{font={scriptsize}}
		\caption*{NBR2NBR}
	\end{subfigure}
	\begin{subfigure}{0.185\linewidth}
		\setlength{\abovecaptionskip}{1pt}
		\centering
		\includegraphics[height=0.6251in]{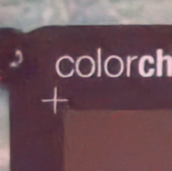}
		\captionsetup{font={scriptsize}}
		\caption*{Blind2Unblind}
		\includegraphics[height=0.6251in]{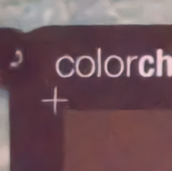}
		\captionsetup{font={scriptsize}}
		\caption*{Ours}
	\end{subfigure}
	\begin{subfigure}{0.41\linewidth}
  		\setlength{\abovecaptionskip}{1pt}
		\centering
		\includegraphics[height=1.375in]{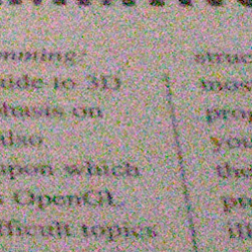}
		\captionsetup{font={scriptsize}}
		\caption*{Noisy}
	\end{subfigure}
	\begin{subfigure}{0.185\linewidth}
		\setlength{\abovecaptionskip}{1pt}
		\centering
		\includegraphics[height=0.6251in]{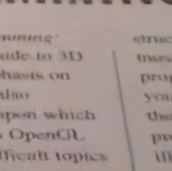}
		\captionsetup{font={scriptsize}}
		\caption*{Baseline, N2C}
		\includegraphics[height=0.6251in]{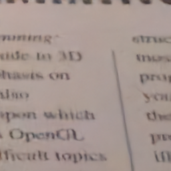}
		\captionsetup{font={scriptsize}}
		\caption*{Baseline, N2N}
	\end{subfigure}
	\begin{subfigure}{0.185\linewidth}
		\setlength{\abovecaptionskip}{1pt}
		\centering
		\includegraphics[height=0.6251in]{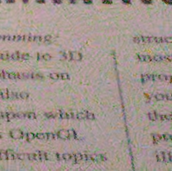}
		\captionsetup{font={scriptsize}}
		\caption*{BM3D}
		\includegraphics[height=0.6251in]{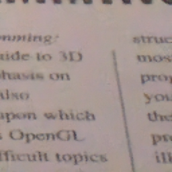}
		\captionsetup{font={scriptsize}}
		\caption*{NBR2NBR}
	\end{subfigure}
	\begin{subfigure}{0.185\linewidth}
		\setlength{\abovecaptionskip}{1pt}
		\centering
		\includegraphics[height=0.6251in]{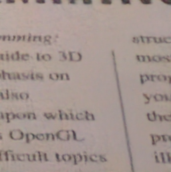}
		\captionsetup{font={scriptsize}}
		\caption*{Blind2Unblind}
		\includegraphics[height=0.6251in]{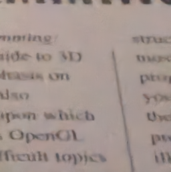}
		\captionsetup{font={scriptsize}}
		\caption*{Ours}
	\end{subfigure}
  \setlength{\abovecaptionskip}{-0.3cm}
  \setlength{\belowcaptionskip}{-0.6cm}
  \caption{Visual comparison of different models denoising for Raw-RGB images in SIDD benchmark. PSNR/SSIM can not be calculated since clean images are not available. The images are well shown from Raw-RGB to sRGB using the official ISP tool \cite {b45}.}
  \label{fig:figure 8}
\end{figure}

\begin{table}[htbp]
  \setlength{\abovecaptionskip}{0cm}
  \setlength{\belowcaptionskip}{0cm}
  \centering
  \caption{Quantitative comparison of denoising results on synthetic datasets in FMDD. DT denotes our Denoised Transformer. The highest PSNR/SSIM among unsupervised denoising methods is highlighted in {\bf bold}, and the second is \underline {underlined}.}
  \label{tab:table 4}
  \begin{tabular*}{\hsize}{l@{\extracolsep{\fill}}l@{\extracolsep{\fill}}c@{\extracolsep{\fill}}c@{\extracolsep{\fill}}c@{\extracolsep{\fill}}}
    \hline
    Methods & Network & \makecell[c]{Confocal \\ Fish} &\makecell[c]{Confocal\\ Mice} & \makecell[c]{Two-Photon\\ Mice} \\
	\hline
	Baseline, N2C\cite{b23}	&U-Net \cite{b23}	&32.79/0.905	&38.40/0.966	&34.02/0.925\\
	Baseline, N2N\cite{b16}	&U-Net \cite{b23}	&32.75/0.903	&38.37/0.965	&33.80/0.923\\
	\hline
	BM3D \cite{b1}	&-	&32.16/0.886	&37.93/0.963 	&33.83/0.924\\
	N2V \cite{b14}	&U-Net \cite{b23}	&32.08/0.886	&37.49/0.960	&33.38/0.916\\
	Laine19-mu (G) \cite{b19}	&U-Net \cite{b23}	&31.62/0.849	&37.54/0.959	&32.91/0.903\\
	Laine19-pme (G) \cite{b19}	&U-Net \cite{b23}	&23.30/0.527 	&31.64/0.881	&25.87/0.418\\
	Laine19-mu (P) \cite{b19}	&U-Net \cite{b23}	&31.59/0.854 	&37.30/0.956	&33.09/0.907\\
	Laine19-pme (P) \cite{b19}	&U-Net \cite{b23}	&25.16/0.597	&37.82/0.959	&31.80/0.820\\
	NBR2NBR \cite{b11}	&U-Net \cite{b23}	&32.11/0.890 	&37.07/0.960	&33.40/0.921\\
	Blind2Unblind\cite{b33}	&U-Net \cite{b23}	&{\bf 32.74/0.897} 	&{\bf 38.44/0.964}	&{\bf 34.03/0.916}\\
	Ours	&DT	&\underline{32.52/0.895}	&\underline{38.21/0.962}	&\underline{33.64/0.914}\\
	\hline
  \end{tabular*}
\vspace{-1.5em}
\end{table}
 Although the method proposed in this paper does not yet outperform the SOTA Blind2Unblind in Raw-RGB and FM, Denoise Transformer has demonstrated its potential to handle complex noise patterns in self-supervised denoising, which is comparable to the performance of SOTA. Fig. \ref{fig:figure 5}, Fig. \ref{fig:figure 8} and Fig. \ref{fig:figure 9} show the visual comparison of SIDD validation, SIDD benchmark and FMDD. There are color shifts and texture loss in NBR2NBR and Blind2Unblind, especially for the top row images in Fig. \ref{fig:figure 5}. Compared to other models, our model has similar performance in general scene images, but has the best denoising performance in low saturated textures and low-light scenes. 
 
\begin{figure}[htbp]
  \setlength{\abovecaptionskip}{0cm}
  \setlength{\belowcaptionskip}{0cm}
	\centering
		\begin{subfigure}{0.192\linewidth}
		\setlength{\abovecaptionskip}{1pt}
		\centering
		\includegraphics[height=0.652in]{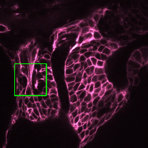}
       \captionsetup{font={scriptsize}}
		\caption*{Original}
	\end{subfigure}
	\begin{subfigure}{0.192\linewidth}
		\setlength{\abovecaptionskip}{1pt}
		\centering
		\includegraphics[height=0.652in]{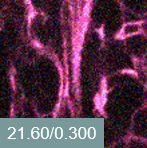}
       \captionsetup{font={scriptsize}}
		\caption*{Noisy}
	\end{subfigure}
	\begin{subfigure}{0.192\linewidth}
		\setlength{\abovecaptionskip}{1pt}
		\centering
		\includegraphics[height=0.652in]{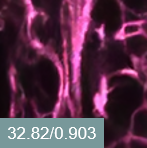}
       \captionsetup{font={scriptsize}}
		\caption*{Baseline, N2C}
	\end{subfigure}
	\begin{subfigure}{0.192\linewidth}
		\setlength{\abovecaptionskip}{1pt}
		\centering
		\includegraphics[height=0.652in]{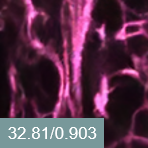}
       \captionsetup{font={scriptsize}}
		\caption*{Baseline, N2N}
	\end{subfigure}
	\begin{subfigure}{0.192\linewidth}
		\setlength{\abovecaptionskip}{1pt}
		\centering
		\includegraphics[height=0.652in]{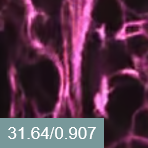}
       \captionsetup{font={scriptsize}}
		\caption*{BM3D}
	\end{subfigure}

 \centering
	\begin{subfigure}{0.192\linewidth}
		\setlength{\abovecaptionskip}{1pt}
		\centering
		\includegraphics[height=0.652in]{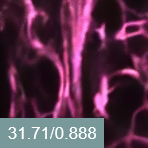}
       \captionsetup{font={scriptsize}}
		\caption*{Laine19-mu(G)}
	\end{subfigure}
	\begin{subfigure}{0.192\linewidth}
		\setlength{\abovecaptionskip}{1pt}
		\centering
		\includegraphics[height=0.652in]{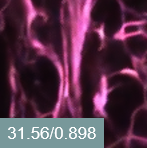}
       \captionsetup{font={scriptsize}}
		\caption*{Laine19-mu(P)}
	\end{subfigure}
	\begin{subfigure}{0.192\linewidth}
		\setlength{\abovecaptionskip}{1pt}
		\centering
		\includegraphics[height=0.652in]{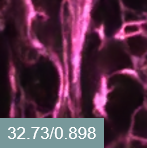}
       \captionsetup{font={scriptsize}}
		\caption*{NBR2NBR}
	\end{subfigure}
	\begin{subfigure}{0.192\linewidth}
		\setlength{\abovecaptionskip}{1pt}
		\centering
		\includegraphics[height=0.652in]{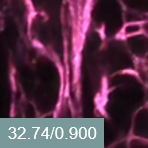}
       \captionsetup{font={scriptsize}}
		\caption*{Blind2Unblind}
	\end{subfigure}
	\begin{subfigure}{0.192\linewidth}
		\setlength{\abovecaptionskip}{1pt}
		\centering
		\includegraphics[height=0.652in]{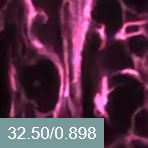}
       \captionsetup{font={scriptsize}}
		\caption*{Ours}
	\end{subfigure}
  \setlength{\abovecaptionskip}{0cm}   
  \setlength{\belowcaptionskip}{-0.3cm}   
  \caption{Visual Comparison of different methods denoising for FM images.}
  \label{fig:figure 9}
\end{figure}
\FloatBarrier
{\bf Computational Complexity Comparison.}  We compute the model size and the FLOPs of state-of-the-art self-supervised image denoising methods that have similar preprocessing and inference flows, as in the following Table \ref {tab:table 5}.

Transformer is known to be computationally intensive. We have made some reductions in the baseline, but it is still more computationally intensive than other models. Given the huge potential of Transformer-based methods in self-supervise denoising, we will try to find a more effective way of combining Transformers with CNN basic modules. At this stage, we can replace some of the CADT units with pure convolution modules to reduce the number of Transformers, which may reduce the amount of computation while maintaining good denoising performance. We believe that our future research on how to reduce the computational complexity of the model is of great importance in order to successfully land in the industry.

\begin{table}[htbp]
  \setlength{\abovecaptionskip}{-0.05cm}
  \setlength{\belowcaptionskip}{-0.2cm}
  \centering
  \caption{Computational Complexity Comparison of several methods.}
  \label{tab:table 5}
  \begin{tabular}{m{2.5cm}m{2cm}<{\centering}m{1.4cm}<{\centering}}
    \hline
    Methods & Parameters (MB) & FLOPs (e9)\\
    \hline
    Blind2Unblind\cite{b33} & 4.4 & 1.17\\
    NBR2NBR\cite{b11}  &  5.0 & 1.37 \\
    Ours &	14 & 3.14\\
    \hline
  \end{tabular}
\vspace{-1em}
\end{table}
\section{Conclusion}
In this paper, in order to improve the local limitations of CNN networks in self-supervised image denoising, we extend the Transformer-based method, inspired by its successful application in computer vision. We propose a basic network, called Denoise Transformer, which is composed of CADTs and SNE to realize self-supervised denoising in two stages.  Each CADT contains a global feature extraction branch and a local feature extraction branch, and the output of the CADTs is our desired noise information. When we subtract the noise from the input through residual learning, we get the denoised output of the first stage . We then design a low computational complexity SNE to be connected after the denoised output to extract its global noise for secondary denoising. Sufficient experiments indicate that abandoning the re-visible branch is suitable for our proposed Denoise Transformer, and both CADTs and SNE contribute a better performance than traditional methods and most self-supervised training methods. Our method demonstrates the strong denoising ability of Transformer under the complex noise pattern and its potential for self-supervised image denoising. We are the first to propose a Transformer-based method for self-supervised image denoising, which fills the gap in the corresponding field and achieves state-of-the-art competitive performance on the public dataset, especially in denoising the noisy images with low-saturated textures and low-light scenes. Although we extend the Transformer based method in new application of self-supervised image denoising, its large computational load is difficult as well as the denosing of high-saturated region. We attribute them to the effectiveness of the combination of Transformer and CNN. In the future, we will investigate the effectiveness of possible parameter reduction schemes and the balance between global and the local features. Given Transformer's strong performance in computer vision, we believe that it has a greater hidden potential for self-supervised image denoising, and hope that the method proposed in this paper can provide more inspiration to researchers.

\EOD

\end{document}